\documentclass[letterpaper, 10pt, conference]{ieeeconf}      

\IEEEoverridecommandlockouts                              
                                                       
\usepackage{url}
\usepackage{comment}
\usepackage{booktabs}
\usepackage{amsmath,amssymb,amsfonts, mathtools}
\usepackage{graphicx}
\usepackage{bm}
\usepackage{epstopdf}
\usepackage{placeins}
\usepackage{bm}
\usepackage{cite}
\let\labelindent\relax
\usepackage[inline]{enumitem}
\usepackage{multicol}
\usepackage{array}
\usepackage{multirow}
\usepackage{tikz}
\usepackage{caption}
\usepackage{subcaption}
\usepackage{tikz}
\usepackage{svg}

\usepackage[linesnumbered,ruled,vlined]{algorithm2e}
\SetKwInput{KwInput}{Input}
\SetKwInput{KwOutput}{Output}

\newcommand{\vect}[1]{\boldsymbol{#1}}

\DeclareMathOperator*{\argmin}{arg\,min}
\DeclareMathOperator*{\argmax}{arg\,max}

\title{\LARGE \bf Parameter-Free Segmentation of Robot Movements with Cross-Correlation Using Different Similarity Metrics}

\author{Wendy Carvalho,$^*$ Meriem Elkoudi,$^*$ Brendan Hertel, and Reza Azadeh
	\thanks{Authors are with the Persistent Autonomy and Robot Learning (PeARL) Lab, University of Massachusetts Lowell, Lowell, MA 01854, USA. Emails: \texttt{ \{wendy\_carvalho, meriem\_elkoudi\, brendan\_hertel\}@student.uml.edu, reza@cs.uml.edu} \newline
    $^*$ Indicates equal contribution.}
 }
 
\begin{document}

\maketitle
\thispagestyle{empty}
\pagestyle{empty}

\begin{abstract}
    Often, robots are asked to execute primitive movements, whether as a single action or in a series of actions representing a larger, more complex task. These movements can be learned in many ways, but a common one is from demonstrations presented to the robot by a teacher. However, these demonstrations are not always simple movements themselves, and complex demonstrations must be broken down, or segmented, into primitive movements. In this work, we present a parameter-free approach to segmentation using techniques inspired by autocorrelation and cross-correlation from signal processing. In cross-correlation, a representative signal is found in some larger, more complex signal by correlating the representative signal with the larger signal. This same idea can be applied to segmenting robot motion and demonstrations, provided with a representative motion primitive. This results in a fast and accurate segmentation, which does not take any parameters. One of the main contributions of this paper is the modification of the cross-correlation process by employing similarity metrics that can capture features specific to robot movements. To validate our framework, we conduct several experiments of complex tasks both in simulation and in real-world. We also evaluate the effectiveness of our segmentation framework by comparing various similarity metrics.
\end{abstract}

\section{Introduction}
\label{sec:intro}

\begin{figure}[t]
  \centering
    \includegraphics[width=0.98\columnwidth]{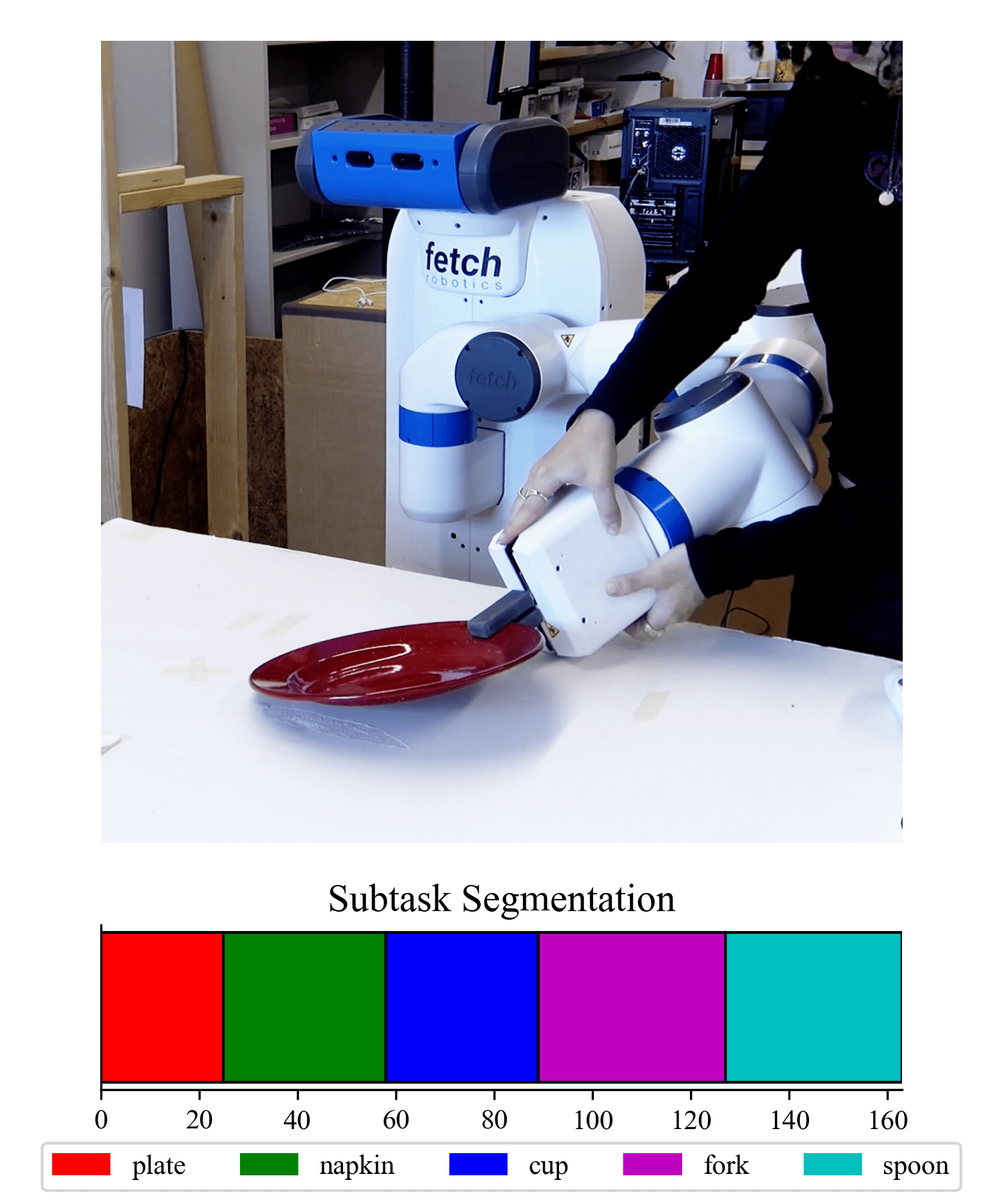}
    \caption{\small{A full-task demonstration of a table setting task, which contains many sub-tasks. The sequencing of sub-tasks is shown.}}
    \label{fig:fig-1}
\end{figure}

As robotic systems are increasingly expected to perform complex, real-world tasks, the ability to autonomously segment a long-duration task into its constituent sub-tasks is crucial for them to be replicated. In many applications, robots obtain new skills through Learning from Demonstration (LfD)~\cite{ravichandar2020recent}, yet the continuous demonstration of a long task often contains multiple sub-tasks that must be accurately identified and segmented for effective learning and reproduction~\cite{kordia2020lgfd, eiband2023segmentation}. One example is the table-setting task shown in Fig.~\ref{fig:fig-1}. To tackle the challenge of robot learning from long and continuous demonstrations, segmentation can be applied to break up the long task into motion primitives (i.e., sub-tasks). However, manually annotating and segmenting demonstrations can be time-consuming and arduous~\cite{chi2024trajectory}. If robots could identify distinct sub-tasks within a more complex task, they would be better able to learn and replicate complex skills~\cite{li2024learning}.

Here, we propose to segment demonstrations using cross-correlation-inspired techniques. In autocorrelation and cross-correlation techniques, a signal is compared either with itself or another signal to find the correlation between the two signals~\cite{rabiner1978theory}. This idea can be applied to trajectory segmentation to find the cross-correlation between motion primitive sub-tasks and a long task. For this, a single long demonstration is compared to several pre-recorded ``representative'' demonstrations of sub-tasks expected to appear within it. However, we consider that there may be variability in these demonstrations, due to differences in velocity, precision, or even environmental conditions. Therefore, a robust similarity metric is essential to perform the correlation effectively. We employ several similarity metrics, including the original cross-correlation function, and compare the results across several environments. While the original cross-correlation function is useful for finding similarity between digital signals, it is not well-suited to finding similarities in robot trajectories. Other similarity metrics proposed here are related to features specific to robot movements, leading to more accurate segmentation. One of the main advantages of this method is that it only requires sub-task and full-task demonstrations, without requiring manual intervention or parameter tuning. Using this segmentation approach, robots can perform complex tasks by applying the detected motion primitives in sequence.

\section{Related Work}

A major challenge in robot task segmentation is the need to manually tune the parameters involved in segmentation algorithms. Approaches such as state clustering require extensive tuning of parameters to work effectively with different tasks~\cite{fan2020hyperparameter}. For example, previous methods enhance the segmentation of tasks by implementing Gaussian Mixture Model (GMM)-based transition clustering~\cite{yamada2023taskseg}. While this improves the reliability of the segmentation by combining visual and kinematic features, the model still relies on intensive tuning of Convolution Neural Networks (CNNs) to incorporate the visual features. In fact, there are several approaches that rely on Deep Learning, which requires many demonstrations~\cite{sorensen2023robot}. Unlike these methods, our approach only requires a single representative demonstration for each sub-task. Other methods do not take representative sub-task demonstrations, but can have a large number of parameters or parameters that may be difficult to tune. In \cite{hertel2023reusable_skills}, a segmentation method is proposed that uses changepoint detection, then probabilistically combines these changepoints across multiple modalities. Each modality can have its own parameters for changepoint detection, resulting in a method that is very difficult to tune, and different tunings leading to very different results. Other methods based on Gaussian Processes and Hidden-Semi-Markov Models (GP-HSMM)~\cite{nakamura2017segmenting} have been proposed, which accurately segment motions. However, this approach also requires the number of classes to be specified beforehand and is computationally expensive, and, therefore, may not be the most suitable for online robot learning. Later, GP-HSMM was extended using a Hierarchical Dirichlet Process to automatically determine the number of classes, known as HDP-GP-HSMM~\cite{nagano2018sequence}. This extension, however, exchanges one hyperparameter (number of classes) with other hyperparameters for a proposed stick-breaking process and Dirichlet process.

\begin{figure*}[t]
  \centering
    \includegraphics[width=0.98\linewidth]{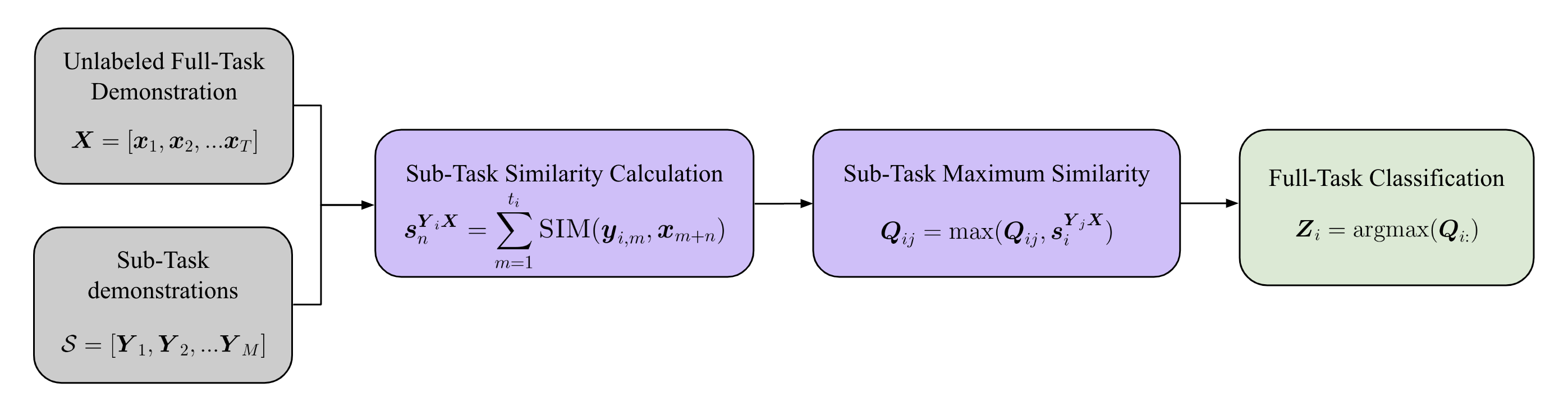}
    \caption{\small{The proposed framework for our parameter-free similarity-based segmentation approach.}}
    \label{fig:framework}
\end{figure*}

A parameter-free approach eliminates the need for manually setting parameters such as the number of classes or CNN hyperparameters. This results in a model that is more flexible and scalable, since it does not rely on fixed parameters that may limit its applicability to new tasks or larger datasets. This is particularly beneficial in robotic applications where task dynamics and sensor noise may vary across different scenarios. By removing the need for frequent retuning, a parameter-free model allows for consistent performance, making it well-suited for LfD frameworks and the transfer of skills between different robots or environments~\cite{argall2009survey}. Additionally, reducing training time enables faster learning and execution of motion tasks. This contributes to the growing body of work aiming to make robotic learning more broadly applicable.

\section{Methodology} 
\label{sec:method}

In this paper, we propose a framework that performs cross-correlation~\cite{box2015time} between a long-duration task and its sub-tasks, shown in Fig.~\ref{fig:framework}.\footnote{ \url{https://github.com/PeARL-robotics/PFCS}} Cross-correlation essentially measures the similarity between observations of a signal at varying points in time. It is useful for detecting patterns and changes in time-series data, making it appropriate for determining transition points between different robot tasks, identifying repetitive patterns, and segmenting said tasks. 

When a robot performs a complex task, the motion of its joints and end-effector follows a structured pattern that consists of repetitive movements. Therefore, by observing how data changes over time during a long task and computing its cross-correlation in relation to specific sub-tasks, we can determine the segments of the time-series data that have the highest similarity to each other. However, there are several challenges in developing correlation segmentation, including
\begin{enumerate*}[label=(\roman*)]
  \item handling variability in demonstrations, and
  \item finding the most appropriate motion features.
\end{enumerate*}
To address these challenges, we apply cross-correlation to analyze temporal variations in the robot's end-effector position for a high-level view of the robot's spatial trajectory. This end-effector position data forms the basis of our time-series input and is used consistently across both full-task and sub-task demonstrations for computing similarity. We employ several different similarity metrics for calculating this correlation, which enables different and more robust features to be captured in segmentation. Although we investigate three different similarity metrics, others could be used. While we use end-effector data in our experiments, this method could be generally applied to joint data or other modalities as well.

Typically, autocorrelation is used to find a signal obscured by noise, performed by correlating a signal with itself~\cite{box2015time}. Essentially, given some time signal $f$, we wish to compute
\begin{equation}
    a^{ff} = f * f,
\end{equation}
where $a^{ff}$ is the autocorrelated signal and $*$ indicates the correlation operator. This autocorrelated signal shows where $f$ shows the most similarity with itself. When the two signals are different, this is known as cross-correlation~\cite{rabiner1978theory}. For example, given some noisy time signal $g$, we can look for $f$ in $g$ by finding where 
\begin{equation}
    c^{fg} = f * g,
\end{equation}
is maximized, where $c^{fg}$ is the cross-correlated signal. This can be applied to robot demonstrations, as full-task demonstrations are given as some time signal $\vect{X} = [\vect{x}_1,\vect{x}_2, \dots, \vect{x}_T] \in \mathbb{R}^{T \times d}$, with individual $d$-dimensional points $\vect{x}_i$. We assume we are also given demonstrations of $M$ sub-tasks in the set $\mathcal{S} = \{ \vect{Y}_1, \vect{Y}_2, \dots, \vect{Y}_M \}$ with an individual sub-task demonstration $\vect{Y}_i = [\vect{y}_{i, 1},\vect{y}_{i, 2}, \dots, \vect{y}_{i, t_i}] \in \mathbb{R}^{t_i \times d}$ and $t_i < T \ \forall \ i=1,\dots,M$. These ``time signals'' could be joint-space trajectories, task-space (end-effector) trajectories, or some other mode. We assume that the set of sub-task demonstrations $\mathcal{S}$ is known in advance. If the sub-tasks are not provided, an additional method for their discovery must be applied prior to segmentation, such as those applied in Visual Imitation Learning (VIL)~\cite{jonnavittula2025view}. We wish to find the cross-correlation between each sub-task demonstration and the full-task demonstration, which will reveal where the sub-task demonstration may appear in the full-task demonstration. For two finite discrete functions, the cross-correlation is defined as 
\begin{equation}
     c^{fg}_n = (f * g)_n = \sum_{m=1}^{N} \overline{f(m)}g(m+n),
\end{equation}
where $\overline{f(m)}$ represents the complex conjugate of $f$~\cite{wang2019kernel}. In the case of robot demonstrations, we only work within the real domain, so $\overline{\vect{X}} = \vect{X}$. Therefore, we apply this to full-tasks and sub-tasks as 
\begin{equation}
    \vect{c}^{\vect{Y}_i \vect{X}}_n = (\vect{Y}_i * \vect{X})_n = \sum_{m=1}^{t_i} \vect{y}_{i, m} \cdot \vect{x}_{m+n},\label{eq:cross-corr}
\end{equation}
where $\vect{c}^{\vect{Y}_i \vect{X}}_n$ is the cross-correlation between sub-task demonstration $\vect{Y}_i$ and the full-task demonstration $\vect{X}$ at some point along the demonstration $n$, and $\cdot$ represents the dot product of two vectors. By doing this for all possible points along the demonstration, we find $\vect{c}^{\vect{Y}_i \vect{X}} = [\vect{c}^{\vect{Y}_i \vect{X}}_0, \vect{c}^{\vect{Y}_i \vect{X}}_1, \dots, \vect{c}^{\vect{Y}_i \vect{X}}_{T-t_i} ]$, which we can use to find the most likely position of the start of sub-task $\vect{Y}_i$ as
\begin{equation}
    P_i = \underset{n}{\argmax} (\vect{c}^{\vect{Y}_i \vect{X}} ), \label{eq:pred}
\end{equation}
where $P_i$ is the predicted start of the sub-task.

While cross-correlation does find the likely position of a sub-task within a full-task demonstration, it does not include the ability to be customized to different robot tasks, which may present certain spatial or geometric features~\cite{Ravichandar2019MCCB}. Therefore, we propose a generalization of \eqref{eq:cross-corr}, where an arbitrary similarity metric is used. Different similarity metrics can lead to different results, depending upon the importance of that metric with respect to the task~\cite{hertel2021SAMLfD}. Cross-correlation with an arbitrary similarity metric takes the form
\begin{equation}
    \vect{s}^{\vect{Y}_i \vect{X}}_n = \sum_{m=1}^{t_i} \text{SIM}(\vect{y}_{i, m}, \vect{x}_{m+n}), \label{eq:sim-corr}
\end{equation}
where $s^{\vect{Y}_i \vect{X}}_n$ is the \textit{similarity} correlation between sub-task demonstration $\vect{Y}_i$ and the full-task demonstration $\vect{X}$ at some point along the demonstration $n$, and $\text{SIM}$ is a similarity metric. From this similarity correlation, the predicted start of the sub-task can be found using the same method as presented in \eqref{eq:pred}.\footnote{here we use higher-is-better similarity metrics which are compatible with $\argmax$, for lower-is-better, $\argmin$ should be used.} In this paper, we investigate three similarity metrics: cross-correlation similarity (CCS), negative Sum of Squared Errors (SSE), and tangent cosine similarity (COS). Cross-correlation similarity is the same as cross-correlation presented in \eqref{eq:cross-corr}, as 
\begin{equation}
    \text{SIM}_{\text{CCS}}(\vect{y}_{i, m}, \vect{x}_{m+n}) = \vect{y}_{i, m} \cdot \vect{x}_{m+n},
\end{equation}
and gives a higher-is-better similarity metric. SSE is defined as 
\begin{equation}
    \text{SIM}_{\text{SSE}}(\vect{y}_{i, m}, \vect{x}_{m+n}) = -||\vect{y}_{i, m} - \vect{x}_{m+n} ||_2^2,
\end{equation}
where $||\cdot||_2$ represents the $L^2$-norm. Normally, using SSE results in a lower-is-better similarity metric, but we apply a negative to result in a higher-is-better similarity metric. Finally, the tangent cosine similarity is defined as 
\begin{equation}
    \text{SIM}_{\text{COS}}(\vect{y}_{i, m}, \vect{x}_{m+n}) = \frac{\vect{a}\cdot\vect{b}}{||\vect{a}||_2 ||\vect{b}||_2},
\end{equation}
where $\vect{a} = \vect{y}_{i+1, m} - \vect{y}_{i, m}$ and $\vect{b} = \vect{x}_{m+n+1} - \vect{x}_{m+n}$, or the tangent of $\vect{y}_{i, m}$ and $\vect{x}_{m+n}$, respectively. This similarity measures the difference in angle between two vectors, which, when comparing the tangent vectors, results in a comparison of shape over the demonstrations.

The result of \eqref{eq:pred} only gives the predicted indices of each sub-task. This does not address possible gaps or overlaps in the predictions, which can be addressed in different ways. For some full-task demonstrations, gaps may be necessary, as not every part of the demonstration belongs to one of the sub-tasks. However, it may be important to fully segment a demonstration, and gaps must be filled~\cite{hertel2023reusable_skills}. Overlaps occur when one sub-task is predicted to start before the previous sub-task has finished. The portion where the overlap occurs could belong to either sub-task, and it should be determined which sub-task is a better fit. To record the best fit for each sub-task, we introduce a sub-task similarity matrix $\vect{Q} \in \mathbb{R}^{T \times M}$, which contains the similarity of a full-task demonstration point $\vect{x}_i$ with sub-task $\vect{Y}_j$ at entry $\vect{Q}_{ij}$. This matrix is iteratively updated starting with the first index of $\vect{X}$, as
\begin{equation}
    \vect{Q}_{ij} = \max (\vect{Q}_{ij}, \vect{s}^{\vect{Y}_j \vect{X}}_i),
\end{equation}
where $\vect{Q}_{ij}$ will hold the maximum similarity for corresponding full-task demonstration points and sub-task demonstrations. This can then be used to find the sub-task class of a point as
\begin{equation}
    \vect{Z}_i = \argmax ( \vect{Q}_{i,:}),
\end{equation}
where $\vect{Q}_{i,:}$ is the $i$-th row of the $\vect{Q}$ matrix. Note that this means sub-task demonstrations given in $\mathcal{S}$ may not necessarily show up in the segmentation, as the segmentation will only choose the best sub-tasks. 

Alternatively, we can allow gaps in the segmentation, leaving some parts of the demonstration unclassified. In order to do this, we assume each sub-task appears at most once in the full-task demonstration. Then, we iteratively classify points in $\vect{Z}$, starting from the sub-task with the highest similarity across all sub-tasks. For example, if the similarity for sub-task $\vect{Y}_i$ (stored as $\vect{Q}_{:,i}$) contains $\max(\vect{Q})$, then we set $\vect{Z}_{j:j+t_i} = i$, where $j = \argmax(\vect{Q}_{:,i})$. This ensures that the most likely sub-task is classified first. Then we perform the same operation for the sub-task with the next-highest maximum value, but take care not to overwrite any classes in $\vect{Z}$ which have already been assigned. This procedure prioritizes sub-tasks with maximum similarity, but allows for gaps in the segmentation. The pseudocode for this process is shown in Algorithm 1. The time complexity of this algorithm is $\mathcal{O}(MT\overline{t_i}d)$, where $\overline{t_i}$ is the average value for all $t_i$ and assumes the $\text{SIM}$ operation is $\mathcal{O}(d)$. Overall, this results in a computationally fast segmentation of the task.

\begin{algorithm}[t]
\DontPrintSemicolon
\small
\label{segmentation_alg3}
  \KwInput{Full-task data $\vect{X}$, Sub-tasks demonstrations $\mathcal{S}$}
  \KwOutput{Sub-task classes $\vect{Z}$}
    $\vect{Q} = [-\infty]_{T \times M}$\;
    \For{$i = 1:M$}{  
        $\vect{s}^{\vect{Y}_i \vect{X}} = [0]$ for $1:T - t_i$\; 
        \For{$j=1:T - t_i$} {  
            $\vect{s}^{\vect{Y}_i \vect{X}}_j \longleftarrow \sum_{m=1}^{t_i} \text{SIM}(\vect{y}_{i, m}, \vect{x}_{m+j}) $\;
            \For{$k=j:j + t_i$} {
                $\vect{Q}_{ki} \longleftarrow \max (\vect{Q}_{ki}, \vect{s}^{\vect{Y}_i \vect{X}}_j)$\;
            }
        }
    }
    $\vect{Z} = [-1]_{T}$\;
    \While{ $\max(\vect{Q}) > -\infty$}{
        $i \longleftarrow \argmax(\max(\vect{Q}, \text{axis}=0))$\;
        $j \longleftarrow \argmax(\vect{Q}_{:,i})$\;
        $k \longleftarrow 0$\;
        \While{$\vect{Z}_{j+k} == -1$ \textbf{and} $k < t_i$ \textbf{and} $j+k < T$}{
            $k \longleftarrow k + 1$\;
        }
        $\vect{Z}_{j:j+k} \longleftarrow i$\;
        $\vect{Q}_{:, i} \longleftarrow -\infty$\;
        $\vect{Q}_{j:j+k, :} \longleftarrow -\infty$\;
    }

\caption{Segmentation via Cross-Correlation with Arbitrary Similarity Metrics}
\end{algorithm}

\section{Experiments}
\label{sec:exps}

In this section, we validate the proposed cross-correlation-based segmentation. We first use our method on a simulated handwriting task, then on a real-world table setting task with a Fetch Robotics Fetch Mobile Manipulator as seen in Fig.~\ref{fig:fig-1}.

\subsection{Segmentation on a Handwriting Task}

\begin{figure}[t]
  \centering
    \includegraphics[width=0.98\columnwidth]{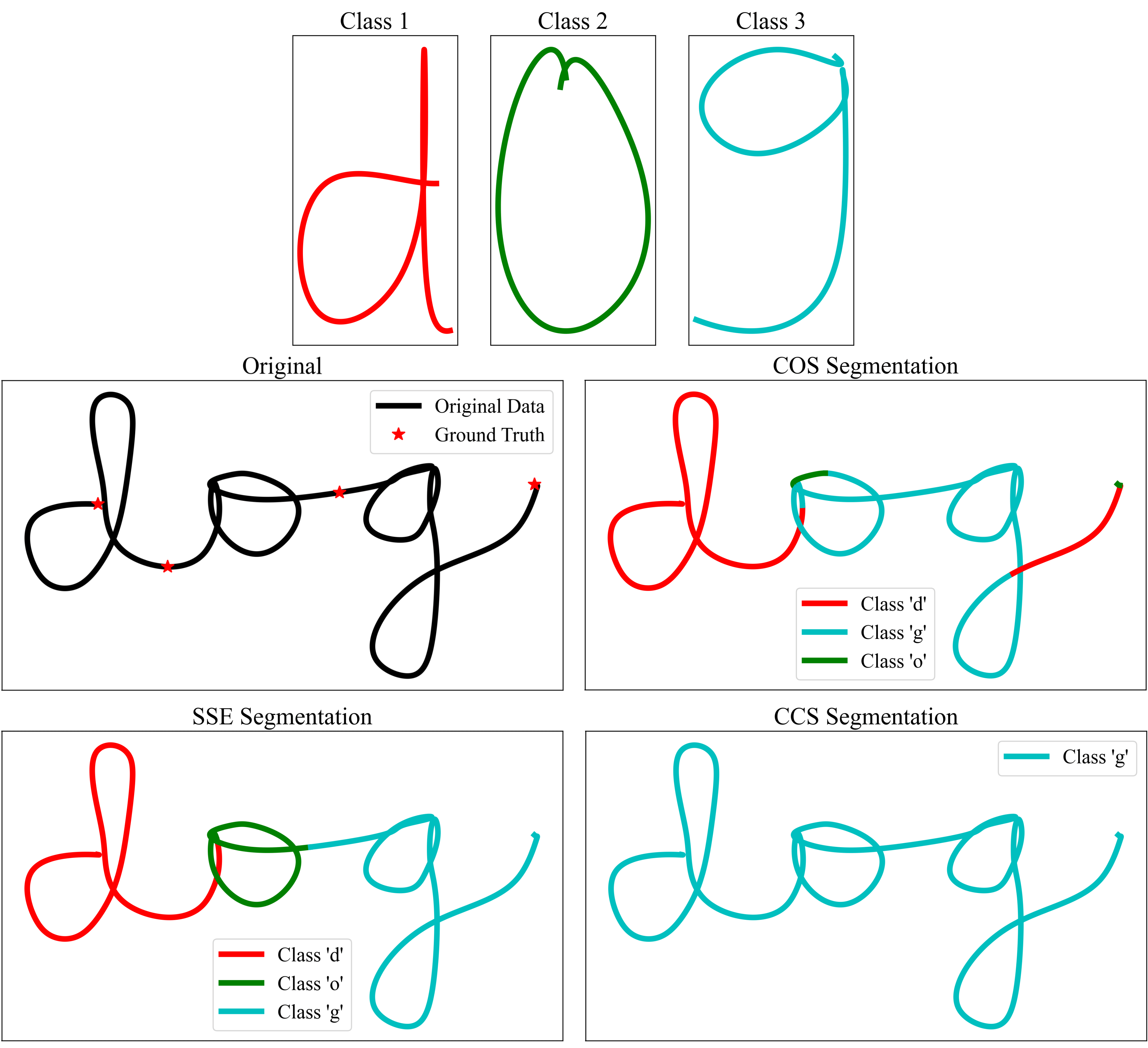}
    \caption{\small{Segmentation results of the handwriting task using COS, SSE, and CCS as similarity metrics. The representative tasks are displayed in the top row.}}
    \label{fig:dog-results}
\end{figure}

We tested the proposed method on a simulated 2D trajectory of the word ``dog'' written in cursive as seen in Fig.~\ref{fig:dog-results}. The goal of the experiment was to segment the trajectory into sub-task movements corresponding to individual letters (``d,'' ``o,'' and ``g'') which were written individually. Here, the full-task demonstration ``dog'' is the continuously written word. The trajectory was captured using a screen capture interface, where a user writes on a screen using a computer mouse. All demonstrations were pre-processed using smoothing and resampling from univariate splines~\cite{dierckx1995curve}.

\begin{table}[t]
    \centering
    \small
    \renewcommand{\arraystretch}{1.2}

    \caption{\small{Accuracy Results for SSE, COS, and CCS Predicting Each Class and Overall Accuracy }}
    \label{tab:accuracy_results}

    \resizebox{\columnwidth}{!}{ 
    \begin{tabular}{@{}c ccc c@{}}
        \toprule
        \textbf{Task} & \textbf{Class 1 (``d'')} & \textbf{Class 2 (``o'')} & \textbf{Class 3 (``g'')} & \textbf{Overall Accuracy} \\ 
        \midrule
        CCS & 0.00\% & 0.00\% & 100.00\% & 40.38\% \\ 
        SSE & 100.00\% & 80.50\% & 100.00\% & 94.80\% \\ 
        COS & 100.00\% & 13.38\% & 85.14\% & 71.25\% \\ 
        \bottomrule
    \end{tabular}
    }
\end{table}

To perform the segmentation, we applied our approach using all three similarity metrics (CCS, SSE, and COS). The results of the segmentation can be seen in Fig.~\ref{fig:dog-results} as well as Table~\ref{tab:accuracy_results}. In Fig.~\ref{fig:dog-results}, the original trajectory is plotted in black, while the detected letter segments are displayed in different colors. Visually, the segmentation using SSE intuitively produces correct results for this experiment, as each letter was properly segmented, with only minor errors observed at the endpoints of some letters. The COS metric performed worse than SSE but better than the CCS metric. COS was able to accurately segment the ``d'' class and only a portion of the ``o'' and ``g'' classes. CCS performed significantly worse than the other similarity metrics, where it predicted the ``g'' class for all data points. This suggests CCS failed to distinguish between the segments effectively. Quantitatively, the accuracy found for each similarity metric further exhibits the discrepancy, as seen in Table~\ref{tab:accuracy_results}. Here, SSE achieved the best segmentation with an overall accuracy score of 94.79\%, followed by COS at 71.25\%, while CCS had a much lower overall accuracy score of 40.38\%. It is also evident in the individual accuracy scores for each sub-task class where SSE had an accuracy of 100\% for class 1 and class 3, and 80.50\% for class 2. In comparison, COS achieved 100\% accuracy for class 1, 13.38\% for class 2 and 85.14\% for class 3. Finally, CCS, which had performed the worst, had accuracy scores of 0\% for both classes 1 and 2, and a 100\% accuracy for class 3 (this is due to classifying the entire trajectory as class 3). Overall, CCS was less suitable for our 2D task, while both SSE and COS were better at identifying the correct sub-task classes.

\begin{figure}[t]
    \centering
    \includegraphics[width=0.98\columnwidth]{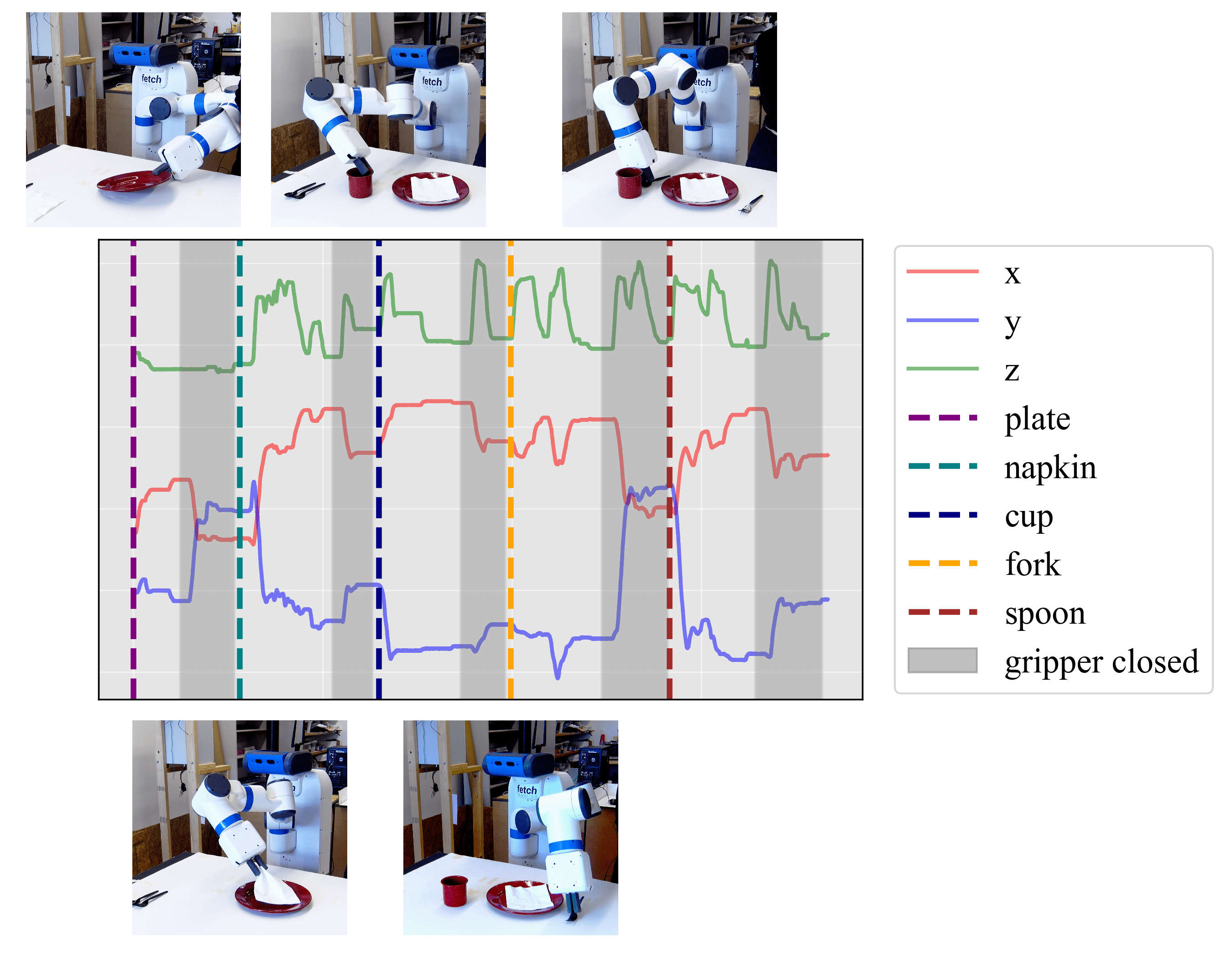}
    \caption{\small{End-effector position and gripper data for full-task with ground truth indices.}}
    \label{fig:ground-truth}
\end{figure}

\subsection{Segmentation of a Table Setting Task}

We also use a real-world table setting task to validate our approach. In this task, several objects must be placed on the table, some of which are ordered (i.e., the plate must be placed before the napkin). First, a demonstration of the table setting task was taken using the Fetch robotic platform, as seen in Fig~\ref{fig:fig-1}. In the demonstration, a plate, a napkin, a cup, a fork, and a spoon were placed in various locations throughout the robot's workspace. Each of these object placements is equivalent to a sub-task in our experiments. The robot is guided using kinesthetic teaching to pick and place each of these objects in the defined sequence to set the table. The demonstrations and the ground truth segments, found manually from human labeling, can be seen in Fig.~\ref{fig:ground-truth}. Here, the end-effector position data used for the experiment ($x$, $y$, and $z$) is shown. The ground truth shows five segments, one corresponding to each object that is picked and placed. For each sub-task, a single demonstration is recorded and used as the representative demo for that sub-task. For this experiment, the segmentation is performed in position space. 

We also experiment with pre-processing the full-task and sub-task demonstration data through smoothing to mitigate noise and outliers. This ensures consistency across the data, and we find that our method performs better under such conditions. It is important to note that although pre-processing techniques are applied to improve the quality of the data, our segmentation does not rely on such techniques. The pre-processing technique involves the selection of parameters that best optimize the data, distinguishing it from our method, which operates without the need for parameter selection. However, pre-processing steps such as smoothing and time alignment are ubiquitous across robotics, and we believe it is important to show results considering such techniques.

We incorporate a smoothing step using the Savitzky-Golay filter\cite{niedzwiecki2021filters}, which is applied separately to each dimension of the data ($x$, $y$, and $z$ of the end-effector position). The filter is applied to both the full-task and sub-task demonstration data before performing the cross-correlation-based segmentation. We empirically determined a window size of 301 and polyorder of 2 were the best parameters for this filter. 

\begin{table*}[ht]
\centering
\caption{\small{Results from segmenting the table setting task, with raw and smoothed data, across all three metrics. For both raw and smoothed data, the COS metric outperforms the other metrics.}}
\label{tab:setting-table}
\begin{tabular}{@{}clrrrrrr@{}}
\toprule
Smoothing &
  \multicolumn{1}{c}{Metric} &
  \multicolumn{1}{c}{Class 1 (plate)} &
  \multicolumn{1}{c}{Class 2 (napkin)} &
  \multicolumn{1}{c}{Class 3 (cup)} &
  \multicolumn{1}{c}{Class 4 (fork)} &
  \multicolumn{1}{c}{Class 5 (spoon)} &
  \multicolumn{1}{c}{Overall Accuracy} \\ \midrule
\multirow{3}{*}{Raw}      & CCS & 0.00\%  & 89.15\%  & 0.00\%  & 72.20\% & 0.00\%  & 34.34\%          \\
                          & SSE & 89.51\% & 26.33\%  & 79.50\% & 83.81\% & 0.00\%  & 53.28\%          \\
                          & COS & 71.82\% & 100.00\% & 0.00\%  & 94.82\% & 95.63\% & \textbf{74.49\%} \\ \midrule
\multirow{3}{*}{Smoothed} & CCS & 0.00\%  & 87.79\%  & 0.00\%  & 71.01\% & 0.00\%  & 33.80\%          \\
                          & SSE & 89.51\% & 26.94\%  & 79.57\% & 83.87\% & 0.00\%  & 53.43\%          \\
                          & COS & 72.62\% & 96.25\%  & 95.05\% & 94.52\% & 95.33\% & \textbf{91.80\%} \\ \bottomrule
\end{tabular}
\end{table*}

\begin{figure}[ht]
    \centering
    \includegraphics[width=0.98\columnwidth]{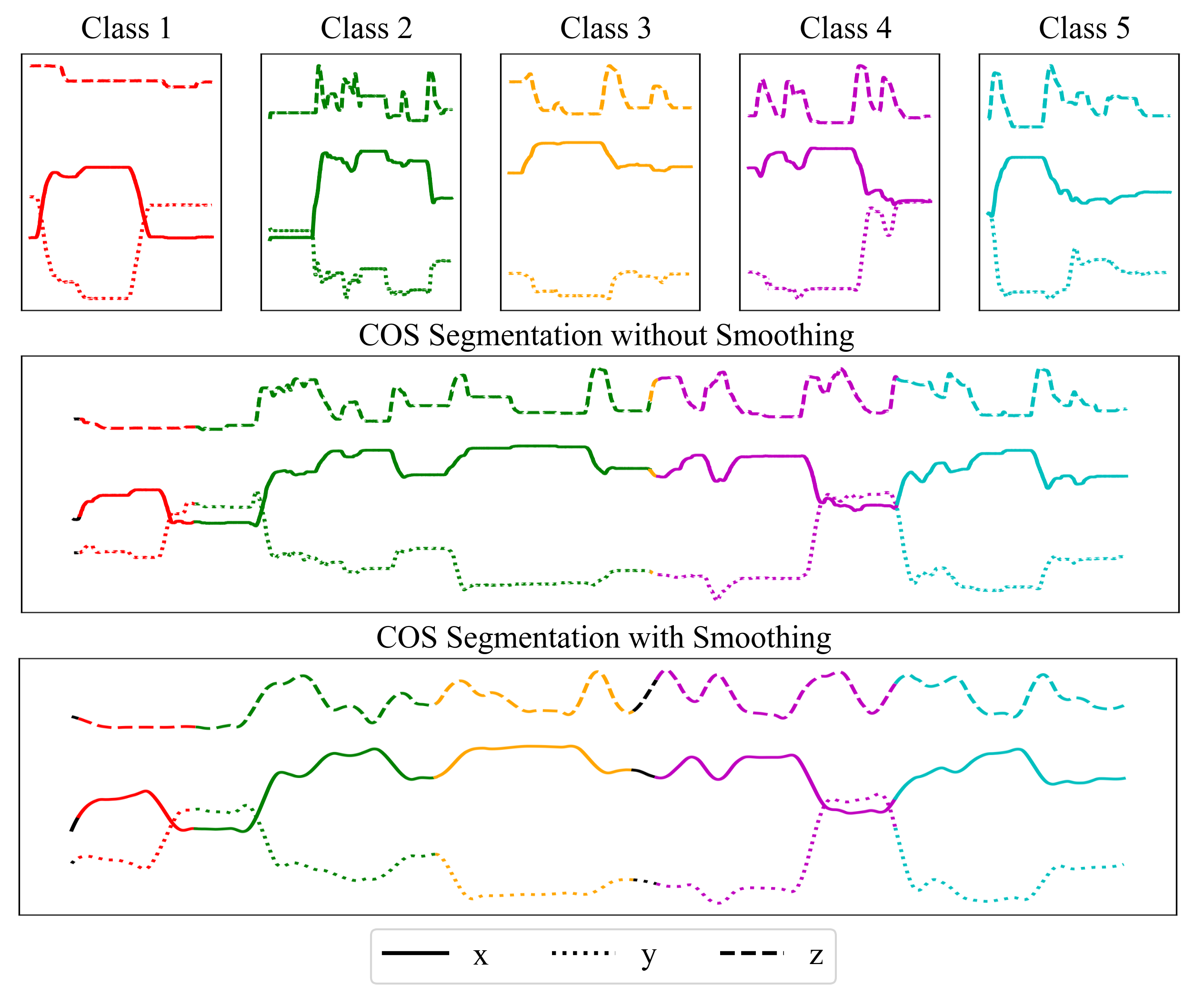}
    \caption{\small{COS Segmentation of the full-task, with the sub-tasks listed in order, with and without smoothing. The representative tasks are displayed in the top row.}}
    \label{fig:task1-cos-segmentation}
\end{figure}

To evaluate the performance of our method, we find the accuracy of the segmentation. In the ground truth, each point is given a true class $\vect{Z}_{true}$. This can be compared with the predicted class $\vect{Z}$ for each point to find the number of accurately predicted points. We also report the prediction accuracy for each sub-task, as some sub-tasks may have higher performance. We report the accuracy for each of the three similarity metrics with and without smoothing in Table~\ref{tab:setting-table}. In both the raw and smoothed cases, CCS leads to the worst results, SSE finds the second-best results, and COS performs the best. CCS is only ever able to accurately detect some of class 2 and class 4, but fails to find the other classes. These results indicate CCS is not useful for this segmentation method. Since CCS is not designed with robotic trajectories in mind, it cannot leverage properties such as spatial, temporal, or angular similarities that may be present in such trajectories. SSE, on the other hand, can detect most of the classes somewhat accurately, but cannot detect class 5 in both the raw and smoothed cases. This could be due to small shifts in position between the sub-task and full-task demonstrations within the environment, which introduce bias in the positional error SSE is designed to measure. Finally, COS leads to the best results for both the raw and smoothed cases, shown in Fig.~\ref{fig:task1-cos-segmentation}. The COS metric relies on finding similarities between the shape of demonstrations, which is consistent across environments. Additionally, the performance of COS is greatly increased with smoothing, as the overall accuracy increases from 74.49\% to 91.80\%. Segmentation with the COS metric performs well across almost all sub-tasks, but there is some error in overwriting of class 1 (plate) early in favor of class 2 (napkin). Additionally, in the raw data case, the segmentation does not detect class 3 (cup), due to temporal differences between demonstrations. This shows how pre-processing can eliminate some errors, leading to better results.

\section{Conclusions \& Future Work}

In this paper, we have proposed a segmentation method inspired by the ideas of autocorrelation and cross-correlation. In correlation, a signal is ``passed over'' another signal to find the similarities or time-shift between the two signals. We apply this idea to find a parameter-free segmentation approach. Additionally, we propose several similarity metrics, which our results indicate perform better than the original cross-correlation similarity metric. We validate our method with a simulated handwriting task as well as a table setting task using a Fetch mobile manipulator in the real-world.

We plan to further refine and extend our approach for robotic sub-task identification in several ways. One possible extension is to improve the similarity measures used. In this work, we propose the use of SSE and COS, which are compared with the original cross-correlation metric. While these similarity metrics improve results, neither is consistently the best metric. Spatial and temporal errors between complex and representative demonstrations can lead to issues in segmentation with these metrics. Finding a metric that can accurately segment all tasks remains a challenge. Further, it may be interesting to develop a way to automatically detect skills that are not present in a sub-task, but are part of the complex task demonstration.

\section*{Acknowledgments}

This research is supported in part by the National Science Foundation (FRR-2237463).

\typeout{}
\bibliographystyle{IEEEtran}
\bibliography{references}

\end{document}